\documentclass[10pt,twocolumn,letterpaper]{article}

\usepackage{cvpr}
\usepackage{times}
\usepackage{epsfig}
\usepackage{graphicx}
\usepackage{float}
\usepackage{amsmath}
\usepackage{amssymb}
\usepackage{soul}
\usepackage{multicol}
\usepackage{blindtext}
\usepackage{here}
\usepackage{caption}
\usepackage{makecell}
\usepackage[normalem]{ulem}
\usepackage{multirow}
\usepackage{booktabs}
\usepackage{color, colortbl}

\definecolor{Gray}{gray}{0.9}
\usepackage[pagebackref=true,breaklinks=true,letterpaper=true,colorlinks,bookmarks=false]{hyperref}

\newcommand{\predTheta}{\hat{\mathbf{\Theta}}}
\newcommand{\predShape}{\hat{\mathbf{\beta}}}
\newcommand{\predPose}{\hat{\mathbf{\theta}}}
	
\newcommand{\gtTheta}{\mathbf{\Theta}}
\newcommand{\gtShape}{\mathbf{\beta}}
\newcommand{\gtPose}{\mathbf{\theta}}

\newcommand{\motionDisc}{\mathcal{D}_M}	
\newcommand{\generator}{\mathcal{G}}

\newcommand{\real}{\mathbb{R}}
\cvprfinalcopy %

\ifcvprfinal\fi

\begin{document}

\title{VIBE: Video Inference for Human Body Pose and Shape Estimation}

\author{%
  Muhammed Kocabas$^{1,2}$,\; Nikos Athanasiou$^1$,\; Michael J. Black$^1$\\\
  \normalsize $^1$Max Planck Institute for Intelligent Systems, T\"{u}bingen, Germany \\
  \normalsize $^2$Max Planck ETH Center for Learning Systems\\
  \normalsize \texttt{\{\href{mailto:mkocabas@tue.mpg.de}{mkocabas},\href{mailto:nathanasiou@tue.mpg.de}{nathanasiou},\href{mailto:black@tue.mpg.de}{black}\}@tue.mpg.de}\\ 
}

\twocolumn[{%
	\renewcommand\twocolumn[1][]{#1}%
	\maketitle
	\begin{center}
		\newcommand{\teaserwidth}{\textwidth}
		\centerline{
			\includegraphics[width=\teaserwidth,clip]{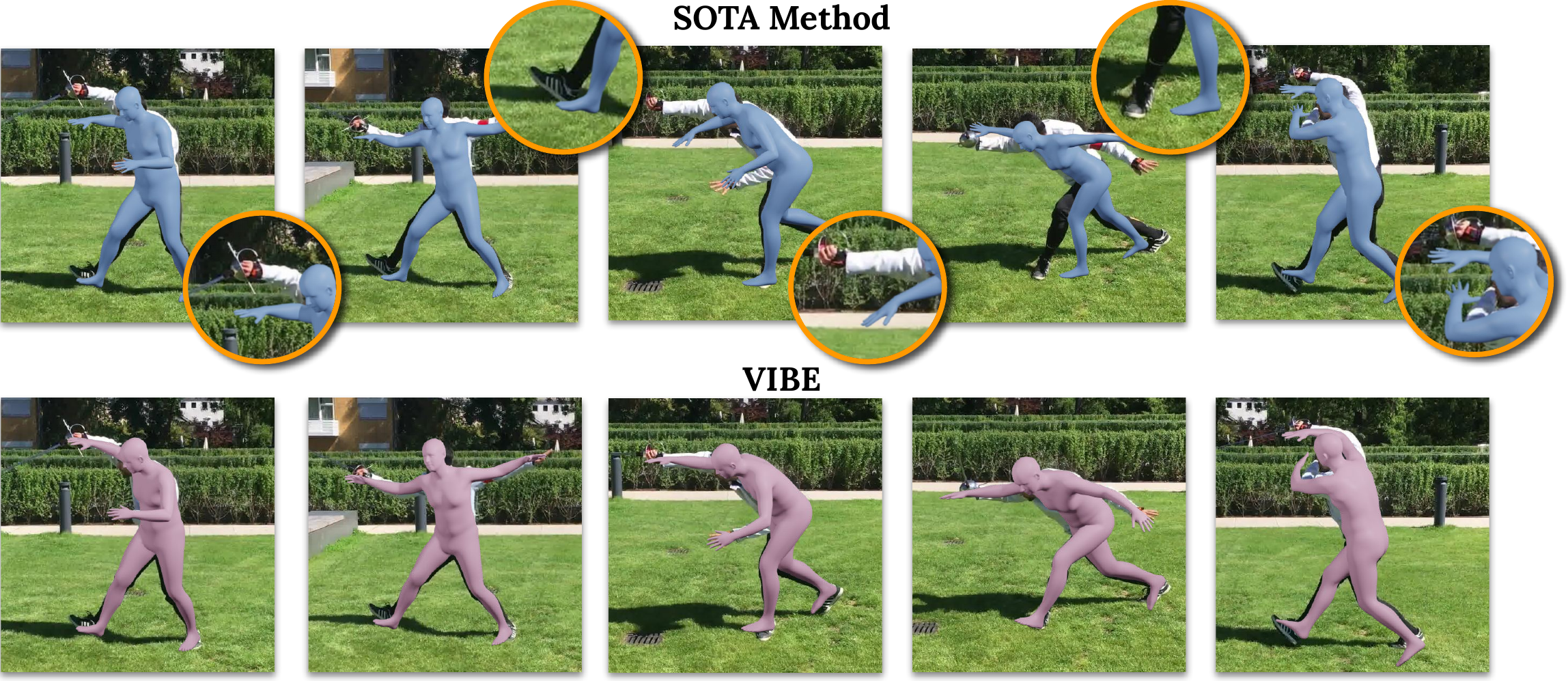}
		}
		\captionof{figure}{Given challenging in-the-wild videos, a recent
			state-of-the-art video-pose-estimation approach~\cite{kanazawa_temporal_hmr}~\textit{(top)}, fails to produce accurate 3D body poses. To address this, we exploit a large-scale motion-capture dataset to train a motion discriminator using an adversarial approach. Our model (VIBE) \textit{(bottom)} is able to produce realistic and accurate pose and shape, outperforming previous work on standard benchmarks.}
		\vspace{-0.05in}
		\label{fig:teaser}
	\end{center}%
}]

\begin{abstract}
Human motion is fundamental to understanding behavior. Despite progress on single-image 3D pose and shape estimation, existing video-based state-of-the-art methods fail to produce accurate and natural motion sequences due to a lack of ground-truth 3D motion data for training. To address this problem, we propose ``\textbf{V}ideo \textbf{I}nference for \textbf{B}ody Pose and Shape \textbf{E}stimation'' (VIBE), which makes use of an existing large-scale motion capture dataset (AMASS) together with unpaired, in-the-wild, 2D keypoint annotations. Our key novelty is an adversarial learning framework that leverages AMASS to discriminate between real human motions and those produced by our temporal pose and shape regression networks. We define a novel temporal network architecture with a self-attention mechanism and show that adversarial training, at the sequence level, produces kinematically plausible motion sequences without in-the-wild ground-truth 3D labels. We perform extensive experimentation to analyze the importance of motion and demonstrate the effectiveness of \textbf{VIBE} on challenging 3D pose estimation datasets, achieving state-of-the-art performance. Code and pretrained models are available at \url{https://github.com/mkocabas/VIBE}
\end{abstract}
\section{Introduction}
\label{introduction}
Tremendous progress has been made on estimating 3D human pose and shape from a single image
 \cite{bogo_smplify,Guler_2019_CVPR,huang_mvsmplify,kanazawa_hmr,kocabas2019epipolar,SPIN:ICCV:2019,lassner_up3d,omran2018nbf,pavlakos2018humanshape}. 
While this is useful for many applications, it is the motion of the body in the world that tells us about human behavior.
As noted by Johansson \cite{johansson_perception} even a few moving point lights on a human body in motion informs us about behavior.
Here we address how to exploit temporal information to more accurately estimate the 3D motion of the body from monocular video.
While this problem has received over 30 years of study, we may ask why reliable methods are still not readily available.
Our insight is that the previous temporal models of human motion have not captured the complexity and variability of real human motions due to insufficient training data.
We address this problem here with a new temporal neural network and training approach, and show that it significantly improves 3D human pose estimation from monocular video.

Existing methods for video pose and shape estimation \cite{kanazawa_temporal_hmr, Sun_2019_ICCV} often fail to produce accurate predictions as illustrated in Fig.~\ref{fig:teaser} (top).
A major reason behind this is the lack of in-the-wild ground-truth 3D annotations, which are non-trivial to obtain even for single images.
Previous work \cite{kanazawa_temporal_hmr, Sun_2019_ICCV} combines indoor 3D datasets with videos having 2D ground-truth or pseudo-ground-truth keypoint annotations. 
However, this has several limitations:
(1) indoor 3D datasets are limited in the number of subjects, range of motions, and image complexity; 
(2) the amount of video labeled with ground-truth 2D pose is still insufficient to train deep networks; and 
(3) pseudo-ground-truth 2D labels are not reliable for modeling 3D human motion. 

To address this, we take inspiration from Kanazawa \etal~\cite{kanazawa_hmr} who train a single-image pose estimator using only 2D keypoints and an {\em unpaired} dataset of {\em static} 3D human shapes and poses using an adversarial training approach.
For video sequences, there already exist in-the-wild videos with 2D keypoint annotations. 
The question is then how to obtain realistic 3D human {\em motions} in sufficient quality for adversarial training.
For that, we leverage the  large-scale 3D motion-capture dataset called AMASS \cite{AMASS:2019}, which is sufficiently rich to learn a model of how people move.
Our approach learns to estimate sequences of 3D body shapes poses from in-the-wild videos such that a discriminator cannot tell the difference between the estimated motions and motions in the AMASS dataset.
As in \cite{kanazawa_hmr}, we also use 3D keypoints when available.

The output of our method is a sequence of pose and shape parameters in the SMPL body model format \cite{looper_smpl}, which is consistent with AMASS and the recent literature.
Our method learns about the richness of how people appear in images and is grounded by AMASS to produce valid human motions. Specifically, we leverage two sources of unpaired information by training a sequence-based generative adversarial network (GAN) \cite{goodfellow_gan}. 
Here, given the video of a person, we train a temporal model to predict the parameters of the SMPL body model for each frame while a motion discriminator tries to distinguish between real and regressed sequences. 
By doing so, the regressor is encouraged to output poses that represent plausible motions through minimizing an adversarial training loss while the discriminator acts as weak supervision. 
The motion discriminator implicitly learns to account for the statics, physics and kinematics of the human body in motion using the ground-truth motion-capture (mocap) data. 
We call our method {\bf VIBE}, which stands for ``\textbf{V}ideo \textbf{I}nference for \textbf{B}ody Pose and Shape \textbf{E}stimation.''

During training, VIBE takes in-the-wild images as input and predicts SMPL body model parameters using a convolutional neural network (CNN) pretrained for single-image body pose and shape estimation \cite{SPIN:ICCV:2019} followed by a temporal encoder and body parameter regressor used in~\cite{kanazawa_hmr}. 
Then, a motion discriminator takes predicted poses along with the poses sampled from the AMASS dataset and outputs a real/fake label for each sequence. 
We implement both the temporal encoder and motion discriminator using Gated Recurrent Units (GRUs)~\cite{gru} to capture the sequential nature of human motion. 
The motion discriminator employs a learned attention mechanism to amplify the contribution of distinctive frames.
The whole model is supervised by an adversarial loss along with regression losses to minimize the error between predicted and ground-truth keypoints, pose, and shape parameters.

At test time, given a video, we use the pretrained CNN \cite{SPIN:ICCV:2019} and our temporal module to predict pose and shape parameters for each frame.   
The method works for video sequences of arbitrary length.
We perform extensive experiments on multiple datasets and outperform all state-of-the-art methods;
see Fig.~\ref{fig:teaser} (bottom) for an example of VIBE's output.
Importantly, we show that our video-based method always outperforms single-frame methods by a significant margin on the challenging 3D pose estimation benchmarks 3DPW~\cite{vonMarcard2018_3dpw} and MPI-INF-3DHP~\cite{mpiiinf3dhp_mono-2017}.
This clearly demonstrates the benefit of using video in 3D pose estimation.

In summary, the key contributions in this paper are:
First, we leverage the AMASS dataset of motions for adversarial training of VIBE.  This encourages the regressor to produce realistic and accurate motions.  %
Second, we employ an attention mechanism in the motion discriminator to weight the contribution of different frames  and show that this improves our results over baselines.
Third, we quantitatively compare different temporal architectures for 3D human motion estimation.
Fourth, we achieve state-of-the-art results on major 3D pose estimation benchmarks. %
Code and pretrained models are available for research purposes at \url{https://github.com/mkocabas/VIBE}.

\begin{figure*}
	\centering
	\includegraphics[width=\textwidth]{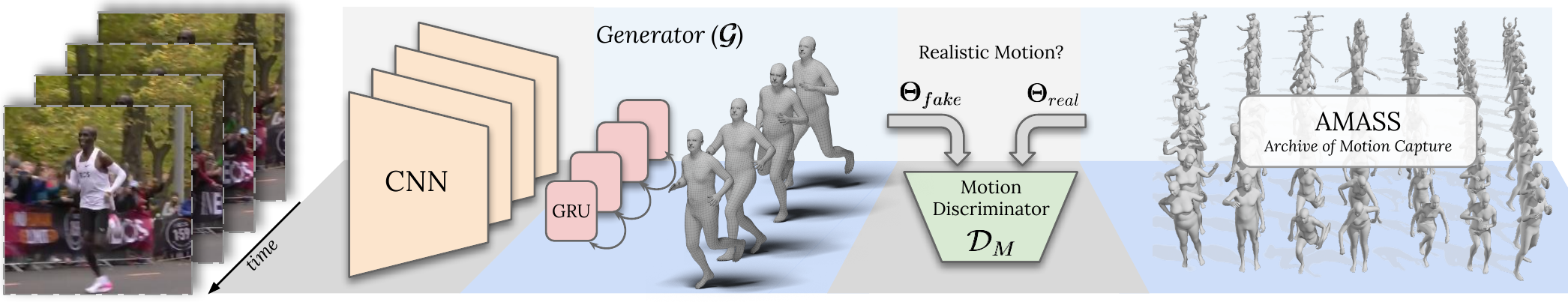}
	\caption{\textbf{VIBE architecture.} VIBE estimates SMPL body
		model parameters for each frame in a video sequence using a
		temporal generation network, which is trained together with a
		motion discriminator. The discriminator has access to a large
		corpus of human motions in SMPL format.}
	\label{fig:model}
\end{figure*}{}

\section{Related Work}
\label{related_work}

{\bf 3D pose and shape from a single image.} 
Parametric 3D human body models \cite{scape,looper_smpl,SMPL-X:2019} are widely used as the output target for human pose estimation because they capture the statistics of human shape and provide a 3D mesh that can be used for many tasks.
Early work explores ``bottom up'' regression approaches, ``top down'' optimization approaches, and multi-camera settings using keypoints and silhouettes as input
 \cite{agarwal2006recovering,balan2008,grauman2003inferring,sigal2008combined}.
These approaches are brittle, require manual intervention, or do not generalize well to images in the wild.
Bogo \etal \cite{bogo_smplify} propose SMPLify, one of the first end-to-end approaches, which fits the SMPL model to the output of a CNN keypoint detector \cite{Leonid2016DeepCut}. Lassner \etal \cite{lassner_up3d} use silhouettes along with keypoints during fitting. Recently, deep neural networks are trained to directly regress the parameters of the SMPL body model from pixels \cite{Guler_2019_CVPR,kanazawa_hmr,omran2018nbf,pavlakos2018humanshape,Tan,tung2017self}. 
Due to the lack of in-the-wild 3D ground-truth labels, these methods use weak supervision signals obtained from a 2D keypoint re-projection loss \cite{kanazawa_hmr,Tan,tung2017self}, use body/part segmentation as an intermediate representation \cite{omran2018nbf,pavlakos2018humanshape}, or employ a human in the loop \cite{lassner_up3d}. Kolotouros \etal~\cite{SPIN:ICCV:2019} combine regression-based and optimization-based methods in a collaborative fashion by using SMPLify in the training loop. At each step of the training, the deep network~\cite{kanazawa_hmr} initializes the SMPLify optimization method that fits the body model to 2D joints, producing an improved fit that is used to supervise the network. Alternatively, several non-parametric body mesh reconstruction methods \cite{kolotouros2019cmr,pifu,varol_bodynet} has been proposed. Varol \etal \cite{varol_bodynet} use voxels as the output body representation. Kolotouros \etal \cite{kolotouros2019cmr} directly regress vertex locations of a template body mesh using graph convolutional networks~\cite{kipf2017semi}. Saito \etal~\cite{pifu} predict body shapes using pixel-aligned implicit functions followed by a mesh reconstruction step. Despite capturing the human body from single images, when applied to video, these methods yield jittery, unstable results.

{\bf 3D pose and shape from video.} 
The capture of human motion from video has a long history.
In early work, Hogg \etal \cite{hogg_walking} fit a simplified human body model to images features of a walking person.
Early approaches also exploit methods like PCA and GPLVMs to learn motion priors from mocap data \cite{ormoneit_cyclic_2001,Urtasun2006} but these approaches were limited to simple motions.
Many of the recent deep learning methods that estimate human pose from video~\cite{dabral2017_tpnet,hossain2018exploiting,Mehta2018XNectRM,pavllo:videopose3d:2019,VNect_SIGGRAPH2017} focus on joint locations only. 
Several methods
\cite{dabral2017_tpnet,hossain2018exploiting,pavllo:videopose3d:2019} use a two-stage approach to ``lift'' off-the-shelf 2D keypoints into 3D joint locations. 
In contrast, Mehta \etal~\cite{Mehta2018XNectRM,VNect_SIGGRAPH2017} employ end-to-end methods to directly regress 3D joint locations. 
Despite impressive performance on indoor datasets like Human3.6M~\cite{ionescu_h36m}, they do not perform well on in-the-wild datasets like 3DPW~\cite{vonMarcard2018_3dpw} and MPI-INF-3DHP~\cite{mpiiinf3dhp_mono-2017}. Several recent methods recover SMPL pose and shape parameters from video by extending SMPLify over time to compute a consistent body shape and smooth motions \cite{arnab_kineticspose,huang_mvsmplify}.
Particularly, Arnab \etal \cite{arnab_kineticspose} show that Internet videos annotated with their version of SMPLify help to improve HMR when used for fine tuning. 
Kanazawa \etal~\cite{kanazawa_temporal_hmr} learn human motion kinematics 
by predicting past and future frames\footnote{Note that they refer to kinematics over time as dynamics.}. 
They also show that Internet videos annotated using a 2D keypoint detector can mitigate the need for the in-the-wild 3D pose labels.
Sun \etal~\cite{Sun_2019_ICCV} propose to use a transformer-based temporal model \cite{vaswani_transformers}  to improve the performance further. They propose an unsupervised adversarial training strategy that learns to order shuffled frames.   

{\bf GANs for sequence modeling.}
Generative adversarial networks GANs~\cite{arjovsky2017wasserstein,goodfellow_gan,isola2017image,liu2017unsupervised} have had a significant impact on image modeling and synthesis. 
Recent works have incorporated GANs into recurrent architectures to model sequence-to-sequence tasks like machine translation~\cite{sutskever2014sequence,wu_adv_nmt,yang-etal-2018-improving}. Research in motion modelling has shown that combining sequential architectures and adversarial training can be used to predict future motion sequences based on previous ones~\cite{BarsoumCVPRW2018,Gui_2018_ECCV} or to generate human motion sequences~\cite{Aksan_2019_ICCV}. 
In contrast, we focus on adversarially refining predicted poses conditioned on the sequential input data.
Following that direction, we employ a motion discriminator that encodes pose and shape parameters in a latent space using a recurrent architecture and an adversarial objective taking advantage of 3D mocap data~\cite{AMASS:2019}.

\section{Approach}
\label{methods}

The overall framework of VIBE is summarized in Fig.~\ref{fig:model}. 
Given an input video $V = \{I_t\}_{t=1}^T$ of length $T$, of a single person, we extract the features of each frame $I_t$ using a pretrained CNN. We train a temporal encoder composed of bidirectional Gated Recurrent Units (GRU) that outputs latent variables containing information incorporated from past and future frames. Then, these features are used to regress the parameters of the SMPL body model at each time instance.

SMPL represents the body pose and shape by $\Theta$, which consists of the pose and shape parameters $\gtPose \in \real^{72}$ and $\gtShape \in \mathbb{R}^{10}$ respectively. 
The pose parameters include the global body rotation and the relative rotation of 23 joints in axis-angle format. 
The shape parameters are the first 10 coefficients of a PCA shape space; here we use the gender-neutral shape model as in previous work \cite{kanazawa_hmr,SPIN:ICCV:2019}
Given these parameters, the SMPL model is a differentiable function, $\mathcal{M}(\theta,\beta) \in \real^{6890 \times 3}$, that outputs a posed 3D mesh.

Given a video sequence, VIBE computes
$\predTheta=[({\predPose_1,\cdots, \predPose_T}), \predShape]$ where $\predPose_t$ are the pose parameters at time step $t$ and $\predShape$ is the single body shape prediction for the sequence. 
Specifically,  for each frame we predict the body shape parameters. Then, we apply average pooling to get a single shape ($\predShape$) across the whole input sequence.
We refer to the model described so far as the temporal generator $\generator$. 
Then, output, $\predTheta$, from $\generator$ and samples from AMASS, $\gtTheta_{real}$, are given to a motion discriminator, $\motionDisc$, in order to differentiate fake and real examples.

\subsection{Temporal Encoder}
The intuition behind using a recurrent architecture is that future frames can benefit from past video pose information. This is useful when the pose of a person is ambiguous or the body is partially occluded in a given frame. Here, past information can help resolve and constrain the pose estimate. The temporal encoder acts as a generator that, given a sequence of frames $I_1,\ldots, I_T$, outputs the corresponding pose and shape parameters in each frame. A sequence of $T$ frames is fed to a convolutional network, $f$, which functions as a feature extractor and outputs a vector $f_i \in \real^{2048}$ for each frame $f(I_1),\ldots, f(I_T)$.
These are sent to a Gated Recurrent Unit (GRU) layer~\cite{gru} that yields a latent feature vector $g_i$ for each frame, $g(f_1),\ldots, g(f_T)$, based on the previous frames.
Then, we use $g_i$ as input to $T$ regressors with iterative feedback as in~\cite{kanazawa_hmr}. The regressor is initialized with the mean pose $\Bar{\Theta}$ and takes as input the current parameters $\Theta_k$ along with the features $g_i$ in each iteration $k$. Following Kolotouros \etal~\cite{SPIN:ICCV:2019}, we use a 6D continuous rotation representation~\cite{Zhou_2019_CVPR} instead of axis angles. 

Overall, the loss of the proposed temporal encoder is composed of 2D ($x$), 3D ($X$), pose ($\theta$) and shape ($\beta$) losses when they are available. This is combined with an adversarial $\motionDisc$ loss. Specifically the total loss of the $\generator$ is: 
\begin{equation}
	L_{\mathcal{G}} = L_{3D} + L_{2D} + L_{SMPL} + L_{adv}
\end{equation}

where each term is calculated as:
\begin{align*}
L_{\mathit{3D}} &=  \sum_{t=1}^{T} \| X_t \; - \; \hat{X}_t \|_2 , \\
L_{\mathit{2D}} &= \sum_{t=1}^{T} \| x_t \; - \; \hat{x}_t \|_2 , \\
L_{\mathit{SMPL}} &= \| \beta \; - \; \hat{\beta} \|_2+ \sum_{t=1}^{T} \| \theta_t \; - \; \hat{\theta}_t \|_2 \;,
\end{align*}
where $L_{adv}$ is the adversarial loss explained below. %

To compute the 2D keypoint loss, we need the SMPL 3D joint locations $\hat{X}(\Theta) = W \mathcal{M}(\theta, \beta)$, which are computed from the body vertices with a pretrained linear regressor, $W$.
We use a weak-perspective camera model with scale and translation parameters $[s,t], t \in \real^2$.
With this we compute the 2D projection of the 3D joints $\hat{X}$, as
$\hat{x} \in \mathbb{R}^{j \times 2} = s\Pi(R\hat{X}(\Theta)) + t $, where $R \in \real^3$ is the global rotation matrix and $\Pi$ represents orthographic projection.

\subsection{Motion Discriminator}
\label{sec:motion_disc}
\begin{figure}[!htbp]
	\centering    
	\includegraphics[width=0.5\textwidth, height= 9cm]{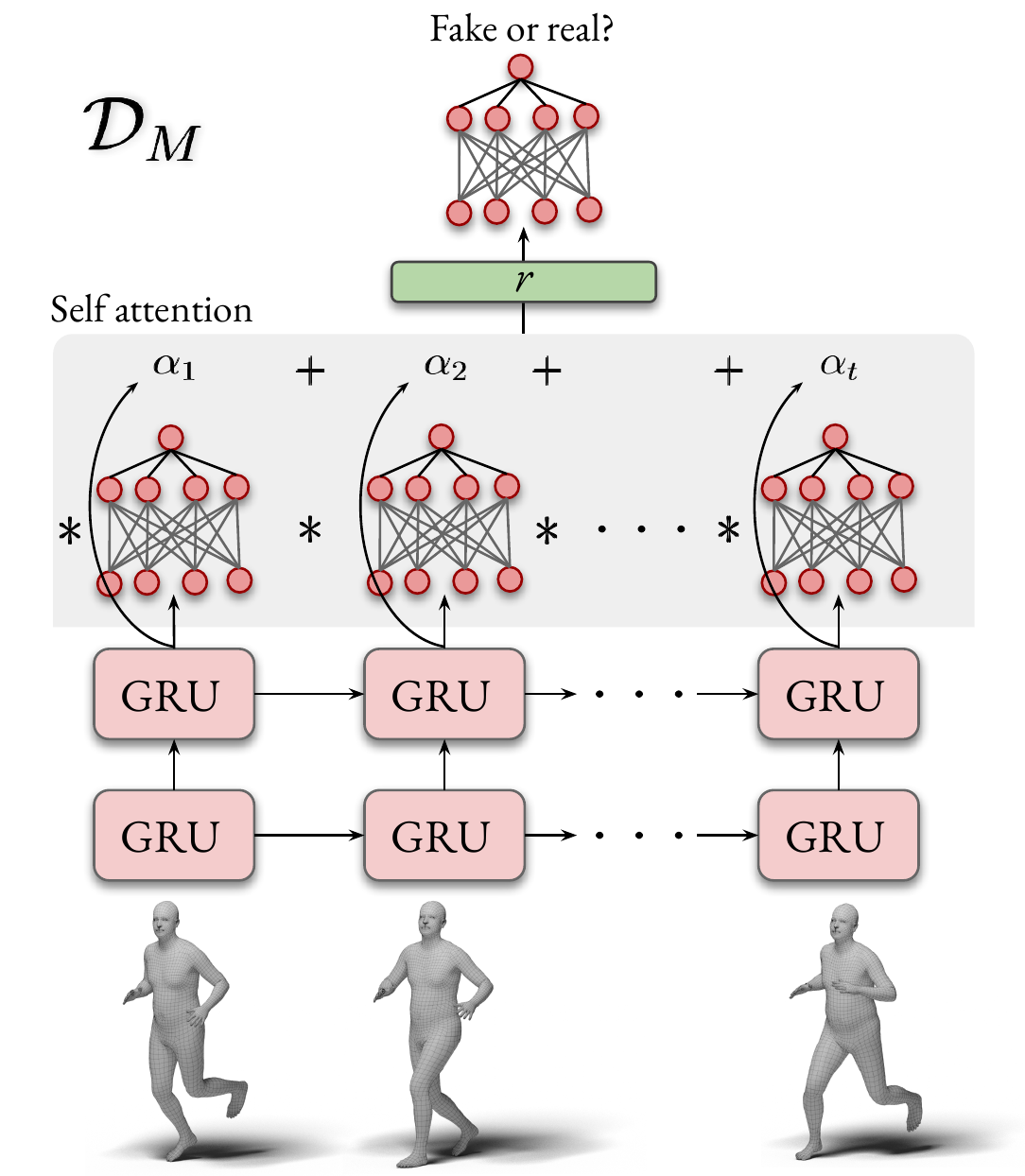}
	\caption{\textbf{Motion discriminator architecture} $\motionDisc$ consists of GRU layers followed by a self attention layer. $\motionDisc$ outputs a real/fake probability for each input sequence.}
	\label{fig:motion}
\end{figure}

The body discriminator and the reprojection loss used in~\cite{kanazawa_hmr} enforce the generator to produce \textit{feasible} real world poses that are aligned with 2D joint locations.  However, single-image constraints are not sufficient to account for sequences of poses. Multiple inaccurate poses may be recognized as valid when the temporal continuity of movement is ignored. To mitigate this, we employ a motion discriminator, $\motionDisc$, to tell whether the generated sequence of poses corresponds to a realistic sequence or not. The output, $\predTheta$, of the generator is given as input to a multi-layer GRU model $f_M$ depicted in Fig.~\ref{fig:motion}, which estimates a latent code $h_i$ at each time step $i$ where $h_i = f_m(\hat{\Theta}_i)$. In order to aggregate hidden states $[h_i, \cdots, h_T]$ we use self attention~\cite{DBLP:journals/corr/BahdanauCB14} elaborated below. Finally, a linear layer predicts a value $\in [0,1]$ representing the probability that $\predTheta$ belongs to the manifold of plausible human motions. The adversarial loss term that is backpropagated to $\generator$ is:  
\begin{equation}
L_{adv} = \mathbb{E}_{\Theta \sim p_{G}}[(\motionDisc(\hat{\mathbf{\Theta}})-1)^2] 
 \label{eq:adv_loss_generator}
\end{equation}
and the objective for $\motionDisc$ is:
\begin{equation}
  L_{\motionDisc} = \mathbb{E}_{\Theta \sim p_{R}}[(\motionDisc(\mathbf{\Theta})-1)^2] + \mathbb{E}_{\Theta \sim p_{G}}[\motionDisc(\hat{\mathbf{\Theta}})^2]
 \label{eq:adv_loss_motion}
\end{equation}
where $p_{R}$ is a real motion sequence from the AMASS dataset, while $p_{G}$ is a generated motion sequence. Since $\motionDisc$ is trained on ground-truth poses, it also learns plausible body pose configurations, hence alleviating the need for a separate single-frame discriminator~\cite{kanazawa_hmr}.
\paragraph{Motion Prior (MPoser).} In addition to the $\motionDisc$, we experiment with a motion prior model, which we call MPoser. It is an extension of the variational body pose prior model VPoser~\cite{SMPL-X:2019} to temporal sequences. We train MPoser as a sequential VAE~\cite{kingma2013autoencoding} on the AMASS dataset to learn a latent representation of plausible human motions. Then, we use MPoser as a regularizer to penalize implausible sequences.  The MPoser encoder and decoder consist of GRU layers that output a latent vector $z_i \in \real^{32}$ for each time step $i$. When we employ MPoser, we disable $\motionDisc$ and add a prior loss $L_{\mathit{MPoser}}=\| z \|_2$ to $L_{\generator}$. 

\paragraph{Self-Attention Mechanism.}
\label{sec:attention}
Recurrent networks update their hidden states as they process input sequentially. As a result, the final hidden state holds a summary of the information in the sequence. We use a self-attention mechanism~\cite{DBLP:journals/corr/BahdanauCB14,baziotis2018ntua} to amplify the contribution of the most important frames in the final representation instead of using either the final hidden state $h_t$ or a hard-choice pooling of the hidden state feature space of the whole sequence. By employing an attention mechanism, the representation $r$ of the input sequence $\hat{\mathbf{\Theta}}$ is a learned convex combination
of the hidden states. The weights $a_i$ are learned by a linear MLP layer $\phi$, and are then normalized using softmax to form a probability distribution. Formally: 
\begin{align}\
\phi_i = \phi(h_i), \quad a_i = \dfrac{\mathrm{e}^{\phi_i}}{\sum_{t=1}^{N} \mathrm{e}^{\phi_t}}, \quad r = \sum_{i=1}^{N} a_i h_i  .
\end{align}
We compare our dynamic feature weighting with a static pooling schema. 
Specifically, the features $h_i$, representing the hidden state at each frame, are averaged and max pooled. Then, those two representations $r_{avg}$ and $r_{max}$ are concatenated to constitute the final static vector, $r$, used for the $D_m$ fake/real  decision.

\subsection{Training Procedure}
We use a ResNet-50 network \cite{he2016resnet} as an image encoder pretrained on single frame pose and shape estimation task~\cite{kanazawa_hmr,SPIN:ICCV:2019} that outputs $f_i \in \mathbb{R}^{2048}$. Similar to ~\cite{kanazawa_temporal_hmr} we precompute each frame's $f_i$ and do not update the ResNet-50. We use $T=16$ as the sequence length with a mini-batch size of 32, which makes it possible to train our model on a single Nvidia RTX2080ti GPU. 
Although, we experimented with $T=[8,16,32,64,128]$, we chose $T=16$, as it yields the best results. 
For the temporal encoder, we use a 2-layer GRU with a hidden size of 1024. The SMPL regressor has 2 fully-connected layers with 1024 neurons each, followed by a final layer that outputs $\predTheta \in \mathbb{R}^{85}$, containing pose, shape, and camera parameters. The outputs of the generator are given as input to the $\motionDisc$ as fake samples along with the ground truth motion sequences as real samples.  The motion discriminator architecture is identical to the temporal encoder. For self attention, we use 2 MLP layers with 1024 neurons each and $tanh$ activation to learn the attention weights. The final linear layer predicts a single fake/real probability for each sample. We also use the Adam optimizer~\cite{Adam} with a learning rate of $5\times10^{-5}$ and $1\times10^{-4}$ for the $\generator$ and $\motionDisc$, respectively. Finally, each term in the loss function has different weighting coefficients. We refer the reader to Sup.~Mat.~for further details.

\begin{table*}[]
	\centering
	\resizebox{\textwidth}{!}{%
		\begin{tabular}{ll|cccc|ccc|cc}
		
			\toprule

			&  & \multicolumn{4}{c}{3DPW} & \multicolumn{3}{c}{MPI-INF-3DHP} & \multicolumn{2}{c}{H36M} \\
			\cmidrule(lr){3-6} \cmidrule(lr){7-9} \cmidrule(lr){10-11}
			& Models & PA-MPJPE $\downarrow$ & MPJPE $\downarrow$ & PVE $\downarrow$ & Accel $\downarrow$ & PA-MPJPE $\downarrow$ & MPJPE $\downarrow$ & PCK $\uparrow$ & PA-MPJPE $\downarrow$ & MPJPE $\downarrow$ \\
			
			\midrule

			\parbox[t]{2mm}{\multirow{6}{*}{\rotatebox[origin=c]{90}{Frame-based}}}  & Kanazawa \etal\cite{kanazawa_hmr} & 76.7 & 130.0 & - & 37.4 & 89.8 & 124.2 & 72.9 & 56.8 & 88 \\
            & \cellcolor{Gray} Omran \etal\cite{omran2018nbf} & \cellcolor{Gray} - & \cellcolor{Gray} - & \cellcolor{Gray} - & \cellcolor{Gray} - & \cellcolor{Gray} - & \cellcolor{Gray} - & \cellcolor{Gray} - & \cellcolor{Gray} 59.9 & \cellcolor{Gray} - \\
			& Pavlakos \etal\cite{pavlakos2018humanshape} & - & - & - & - & - & - & - & 75.9 & - \\
			& \cellcolor{Gray} Kolotouros \etal\cite{kolotouros2019cmr} & \cellcolor{Gray} 70.2 & \cellcolor{Gray} - & \cellcolor{Gray} - & \cellcolor{Gray} - & \cellcolor{Gray} - & \cellcolor{Gray} - & \cellcolor{Gray} - & \cellcolor{Gray} 50.1 & \cellcolor{Gray} - \\
			& Arnab \etal\cite{arnab_kineticspose} & 72.2 & - & - & - & - & - & - & 54.3 & 77.8 \\
			& \cellcolor{Gray} Kolotouros \etal\cite{SPIN:ICCV:2019} & \cellcolor{Gray} 59.2 & \cellcolor{Gray} 96.9 & \cellcolor{Gray} 116.4 & \cellcolor{Gray} 29.8 & \cellcolor{Gray} 67.5 & \cellcolor{Gray} 105.2 & \cellcolor{Gray} 76.4 & \cellcolor{Gray} \textbf{41.1} & \cellcolor{Gray} - \\
			
			\midrule
			
		    \parbox[t]{2mm}{\multirow{5}{*}{\rotatebox[origin=c]{90}{Temporal}}} & Kanazawa \etal\cite{kanazawa_temporal_hmr} & 72.6 & 116.5 & 139.3 & \textbf{15.2} & - & - & - & 56.9 & - \\
			& \cellcolor{Gray} Doersch \etal\cite{doersch_sim2real} & \cellcolor{Gray} 74.7 & \cellcolor{Gray} - & \cellcolor{Gray} - & \cellcolor{Gray} - & \cellcolor{Gray} - & \cellcolor{Gray} - & \cellcolor{Gray} - & \cellcolor{Gray} - & \cellcolor{Gray} - \\
			& Sun \etal\cite{Sun_2019_ICCV} & 69.5 & - & - & - & - & - & - & 42.4 & \textbf{59.1} \\
			\cmidrule(lr){2-11}
			& \cellcolor{Gray} VIBE (direct comp.) & \cellcolor{Gray} 56.5 & \cellcolor{Gray} 93.5 & \cellcolor{Gray} 113.4 & \cellcolor{Gray} 27.1 & \cellcolor{Gray} \textbf{63.4} & \cellcolor{Gray} 97.7 & \cellcolor{Gray} \textbf{89.0} & \cellcolor{Gray} 41.5 & \cellcolor{Gray} 65.9 \\
			& VIBE & \textbf{51.9} & \textbf{82.9} & \textbf{99.1} & 23.4 & 64.6 & \textbf{96.6} & 89.3 & 41.4 & 65.6 \\ 
			
			\bottomrule
			
		\end{tabular}%
	}
	\caption{\textbf{Evaluation of state-of-the-art models on 3DPW, MPI-INF-3DHP, and Human3.6M datasets.} 
	VIBE (direct comp.) is our proposed model trained on video datasets similar to~\cite{kanazawa_temporal_hmr,Sun_2019_ICCV}, while VIBE is trained with extra data from the 3DPW training set. VIBE outperforms all state-of-the-art models including SPIN~\cite{SPIN:ICCV:2019} on the challenging in-the-wild datasets (3DPW and MPI-INF-3DHP)  and obtains comparable result on Human3.6M. ``$-$" shows the results that are not available.}
	\label{tab:sota}
\end{table*}{}

\section{Experiments}
\label{experiments}
We first describe the datasets used for training and evaluation. Next, we compare our results with previous frame-based and video-based state-of-the-art approaches. We also conduct ablation experiments to show the effect of our contributions. Finally, we present qualitative results in Fig.~\ref{fig:qual}. 

\paragraph{Training.} Following previous work~\cite{kanazawa_hmr,kanazawa_temporal_hmr, SPIN:ICCV:2019}, we use batches of mixed 2D and 3D datasets. PennAction~\cite{pennaction} and PoseTrack~\cite{PoseTrack} are the only ground-truth 2D video datasets we use, while InstaVariety~\cite{kanazawa_temporal_hmr} and Kinetics-400~\cite{kinetics400} are pseudo ground-truth datasets annotated using a 2D keypoint detector~\cite{cao2018openpose,kocabas18prn}. 
For 3D annotations, we employ 3D joint labels from MPI-INF-3DHP~\cite{mpiiinf3dhp_mono-2017} and Human3.6M~\cite{ionescu_h36m}.
When used, 3DPW and Human3.6M provide SMPL parameters that we use to calculate $L_{\mathit{SMPL}}$. 
AMASS~\cite{AMASS:2019} is used for adversarial training to obtain real samples of 3D human motion. 
We also use the 3DPW~\cite{vonMarcard2018_3dpw} training set to perform ablation experiments; this demonstrate the strength of our model on  in-the-wild data. 

\paragraph{Evaluation.} For evaluation, we use 3DPW~\cite{vonMarcard2018_3dpw}, MPI-INF-3DHP~\cite{mpiiinf3dhp_mono-2017}, and Human3.6M~\cite{ionescu_h36m}. We report results with and without the 3DPW training to enable direct comparison with previous work that does not use 3DPW for training. 
We report Procrustes-aligned mean per joint position error (PA-MPJPE), mean per joint position error (MPJPE), Percentage of Correct Keypoints (PCK) and Per Vertex Error (PVE).  
We compare VIBE with state-of-the-art single-image and temporal methods. 
For 3DPW, we report acceleration error ($mm/s^2$), calculated as the difference in acceleration between the ground-truth and predicted 3D joints. 

\subsection{Comparison to state-of-the-art results}

Table~\ref{tab:sota} compares VIBE with previous state-of-the-art  frame-based and temporal methods. VIBE (direct comp.) corresponds to our model trained using the same datasets as Temporal-HMR~\cite{kanazawa_temporal_hmr}, while VIBE also uses the 3DPW training set. As standard practice, previous methods do not use 3DPW, however we want to demonstrate that using 3DPW for training improves in-the-wild performance of our model. Our models in Table~\ref{tab:sota} use pretrained HMR from SPIN~\cite{SPIN:ICCV:2019} as a feature extractor. We observe that our method improves the results of SPIN, which is the previous state-of-the-art. Furthermore, VIBE outperforms all previous frame-based and temporal methods on the challenging in-the-wild 3DPW and MPI-INF-3DHP datasets by a significant amount, while achieving results on-par with SPIN on Human3.6M. Note that, Human3.6M is an indoor dataset with a limited number of subjects and minimal background variation, while 3DPW and MPI-INF-3DHP contain challenging in-the-wild videos. 

\begin{figure*}
	\centering
	\includegraphics[width=0.94\textwidth]{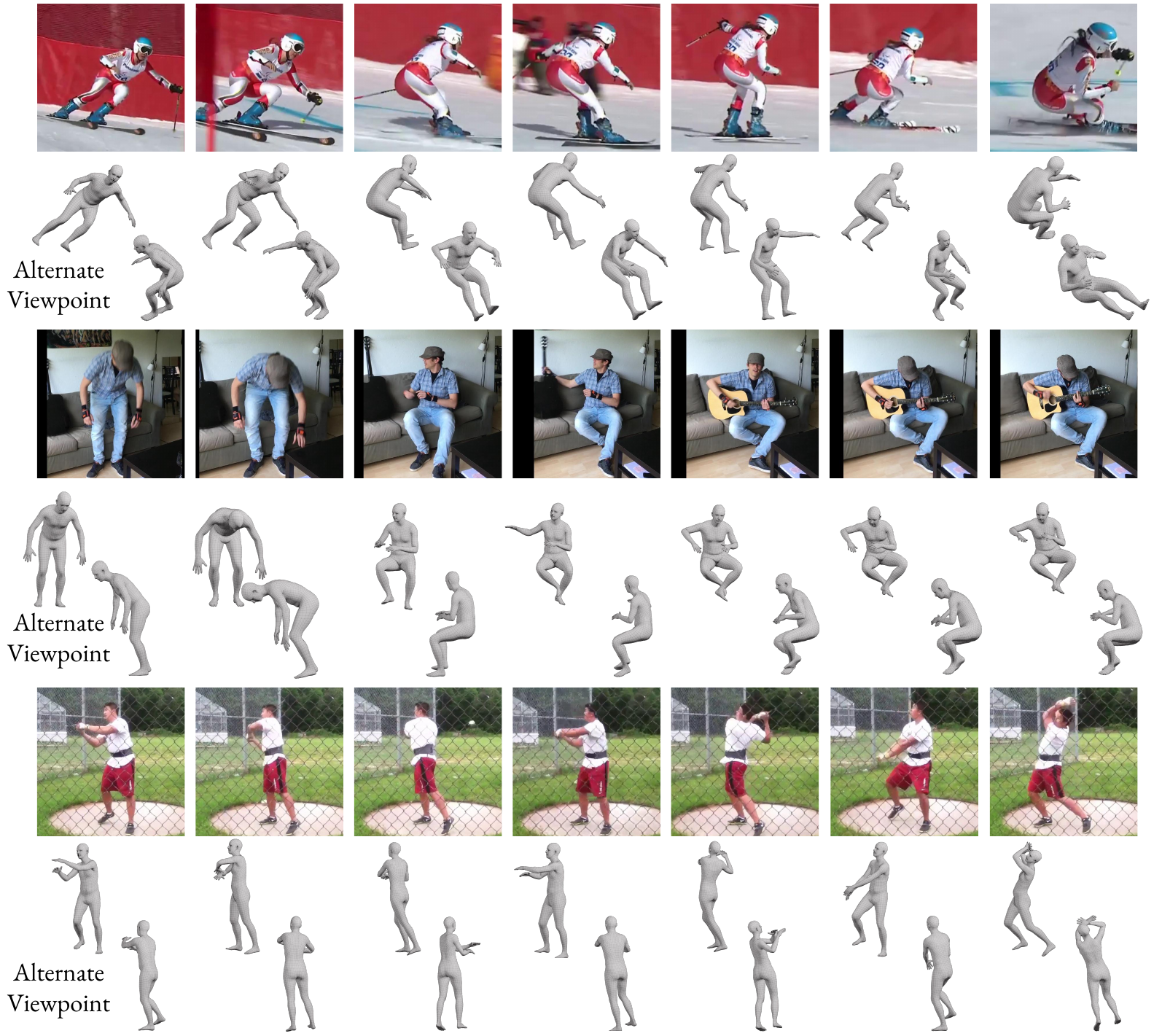}
	\caption{\textbf{Qualitative results of VIBE on challenging in-the-wild sequences.} For each video, the top row shows some cropped images, the middle rows show the predicted body mesh from the camera view, and the bottom row shows the predicted mesh from an alternate view point.}
	\label{fig:qual}
\end{figure*}

\begin{figure*}
	\centering
	\includegraphics[width=\textwidth]{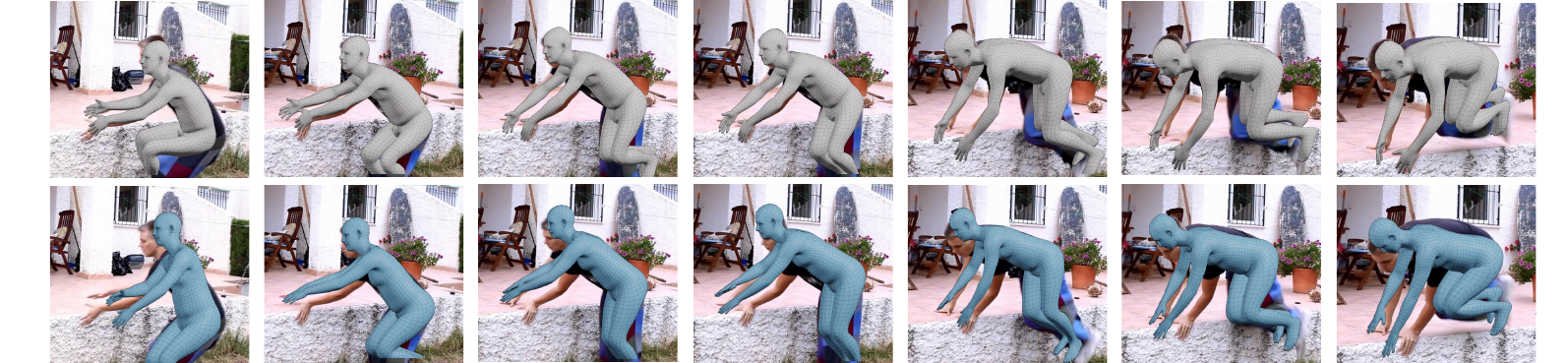}
	\caption{\textbf{Qualitative comparison between VIBE (\textit{top}) and Temporal-HMR~\cite{kanazawa_temporal_hmr} (\textit{bottom}).}  This challenging video contains fast motion, extreme poses, and self occlusion. VIBE produces more accurate poses than Temporal HMR.}           	
	\label{fig:qual_2}
\end{figure*}

We observe  significant improvements in the MPJPE and PVE metrics since our model encourages temporal pose and shape consistency. These results validate our hypothesis that the exploitation of human motion is important for improving pose and shape estimation from video. In addition to the reconstruction metrics, \eg MPJPE, PA-MPJPE, we also report acceleration error (Table~\ref{tab:sota}).
While we achieve smoother results compared with the baseline frame-based methods~\cite{kanazawa_hmr,SPIN:ICCV:2019}, Temporal-HMR~\cite{kanazawa_temporal_hmr} yields even smoother predictions. 
However, we note that Temporal-HMR applies aggressive smoothing that results in poor accuracy on videos with fast motion or extreme poses. 
There is a trade-off between accuracy and smoothness.
We demonstrate this finding in a qualitative comparison between VIBE and Temporal-HMR in Fig.~\ref{fig:qual_2}. 
This figure depicts how Temporal-HMR over-smooths the pose predictions while sacrificing accuracy. 
Visualizations from alternative viewpoints in Fig.~\ref{fig:qual}  show that our model is able to recover the correct global body rotation, which is a significant problem for previous methods.  
This is further quantitatively demonstrated by the improvements in the MPJPE and PVE errors. For video results see the GitHub page.

\begin{table}[]
	\centering
	\resizebox{0.48\textwidth}{!}{%
		\begin{tabular}{l|cccc}
			\toprule
			
			& \multicolumn{4}{c}{3DPW} \\ %
			\midrule
			& PA-MPJPE $\downarrow$ & MPJPE $\downarrow$ & PVE $\downarrow$ & Accel $\downarrow$ \\
			\midrule
			Kanazawa \etal\cite{kanazawa_hmr} & 73.6 & 120.1 & 142.7 & 34.3 \\
			\rowcolor{Gray}
			Baseline (only $\generator$) & 75.8 & 126.1 & 147.5 & 28.3 \\
			$\generator$ + $\motionDisc$ & \textbf{72.4} & \textbf{116.7} & \textbf{132.4} & \textbf{27.8} \\
			
			\midrule
			Kolotouros \etal\cite{SPIN:ICCV:2019} & 60.1 & 102.4 & 129.2 & 29.2 \\
			\rowcolor{Gray}
			Baseline (only $\generator$) & 56.9 & 90.2 & 109.5 & 28.0 \\ %
			$\generator$ + MPoser Prior & 54.1 & 87.0 & 103.9 & 28.2 \\ %
			\rowcolor{Gray}
			$\generator$ + $\motionDisc$ (VIBE) & \textbf{51.9} & \textbf{82.9} & \textbf{99.1} & \textbf{23.4} \\ %
			\bottomrule
		\end{tabular}
	}
	\caption{\textbf{Ablation experiments with motion discriminator $\motionDisc$.} We experiment with several models using HMR~\cite{kanazawa_hmr} and SPIN~\cite{SPIN:ICCV:2019} as pretrained feature extractors and add our temporal generator $\generator$ along with $\motionDisc$. $\motionDisc$ provides consistent improvements over all baselines.}
	\label{tab:ablation}
\end{table}{}

\subsection{Ablation Experiments}
Table~\ref{tab:ablation} shows the performance of models with and without the motion discriminator, $\motionDisc$. 
First, we use the original HMR model proposed by~\cite{kanazawa_hmr} as a feature extractor. Once we add our generator, $\generator$, we obtain slightly worse but smoother results than the frame-based model due to lack of sufficient video training data. This effect has also been observed in the Temporal-HMR method~\cite{kanazawa_temporal_hmr}. Using $\motionDisc$ helps to improve the performance of $\generator$ while yielding smoother predictions. 

When we use the pretrained HMR from~\cite{SPIN:ICCV:2019}, we observe a similar boost when using $\motionDisc$ over using only $\generator$. We also experimented with MPoser as a strong baseline against $\motionDisc$. MPoser acts as a regularizer in the loss function to ensure valid pose sequence predictions. Even though, MPoser performs better than using only $\generator$, it is worse than using $\motionDisc$. One intuitive explanation for this is that, even though AMASS is the largest mocap dataset available, it fails to cover all possible human motions occurring in in-the-wild videos. VAEs, due to over-regularization attributed to the KL divergence term~\cite{tolstikhin2017}, fail to capture real motions that are poorly represented in AMASS.
In contrast, GANs do not suffer from this problem~\cite{Ghoshetal19}. 
Note that, when trained on AMASS, MPoser gives 4.5mm PVE on a held out test set, while the frame-based  VPoser gives 6.0mm PVE error; thus modeling motion matters. 
Overall, results shown in Table~\ref{tab:sota} demonstrate that introducing $\motionDisc$ improves performance in all cases. Although one may think that the motion discriminator might emphasize on motion smoothness over single pose correctness, our experiments with a pose only, motion only, and both modules revealed that the motion discriminator is capable of refining single poses while producing smooth motion.

Dynamic feature aggregation in $\motionDisc$ significantly improves the final results compared to static pooling ($\motionDisc$ - concat), as demonstrated in Table~\ref{tab:attention}. The self-attention mechanism enables $\motionDisc$ to learn how the frames correlate temporally instead of hard-pooling their features. In most of the cases, the use of self attention yields better results. Even with an MLP hidden size of $512$, adding one more layer outperforms static aggregation.
The attention mechanism is able to produce better results because it can learn a better representation of the motion sequence by weighting features from each individual frame. 
In contrast, average and max pooling the features produces a rough representation of the sequence without considering each frame in detail. 
Self-attention involves learning a coefficient for each frame to re-weight its contribution in the final vector ($r$) producing a more fine-grained output. That validates our intuition that attention is helpful for modeling temporal dependencies in human motion sequences.

\begin{table}[]
    \centering
    \resizebox{0.48\textwidth}{!}{%
    \begin{tabular}{l|cccc}
    \toprule
    Model & PA-MPJPE $\downarrow$ & MPJPE $\downarrow$ \\
    
    \midrule
    $\motionDisc$ - concat & 53.7 & 85. 9 \\
    \rowcolor{Gray} 
    $\motionDisc$ - attention [2 layers,512 nodes] & 54.2 & 86.6 \\
    $\motionDisc$ - attention [2 layers,1024 nodes] & \textbf{51.9} & \textbf{82.9} \\
    \rowcolor{Gray} 
    $\motionDisc$  - attention [3 layers,512 nodes] & 53.6 & 85.3 \\
    $\motionDisc$ - attention [3 
    layers,1024 nodes] & 52.4 & 82.7\\
    \bottomrule
    \end{tabular}
}
    \caption{\textbf{Ablation experiments on self-attention.} We experiment with several self-attention configurations and compare our method to a static pooling approach. We report results on the 3DPW dataset with different hidden sizes and numbers of layers of the MLP network.}
    \label{tab:attention}
\end{table}{}

{}

{}
\section{Conclusion}
\label{conclusion}

While current 3D human pose methods work well, most are not trained to estimate human motion in video.
Such motion is critical for understanding human behavior.
Here we explore several novel methods to extend static methods to video: 
(1) we introduce a recurrent architecture that propagates information over time;
(2) we introduce discriminative training of motion sequences using the AMASS dataset;
(3) we introduce self-attention in the discriminator so that it learns to focus on the important temporal structure of human motion;
(4) we also learn a new motion prior (MPoser) from AMASS and show it also helps training but is less powerful than the discriminator.
We carefully evaluate our contributions in ablation studies and show how each choice contributes to our state-of-the-art performance on video benchmark datasets.
This provides definitive evidence for the value of training from video.

Future work should explore using video for supervising single-frame methods by fine tuning the HMR features, examine whether dense motion cues (optical flow) could help, use motion to disambiguate the multi-person case, and exploit motion to track through occlusion. In addition, we aim to experiment with other attentional encoding techniques such as transformers to better estimate body kinematics.

\small{\noindent
{\bf Acknowledgements:} 
We thank Joachim Tesch for helping with Blender rendering. We thank all Perceiving Systems department members for their feedback and the fruitful discussions. This research was partially supported by the Max Planck ETH Center for Learning Systems and the Max Planck Graduate Center for Computer and Information Science.

\noindent
{\bf Disclosure:} 
MJB has received research gift funds from Intel, Nvidia, Adobe, Facebook, and Amazon. 
While MJB is a part-time employee of Amazon, his research was performed solely at, and funded solely by, MPI. 
MJB has financial interests in Amazon and Meshcapade GmbH.}

{\small
\bibliographystyle{ieee_fullname}
\bibliography{references}
}

\newpage

\section{Appendix -- Supplmentary Material}

\subsection{Implementation Details}

\begin{figure}[h]
	\centering    
	\includegraphics[width=0.5\textwidth]{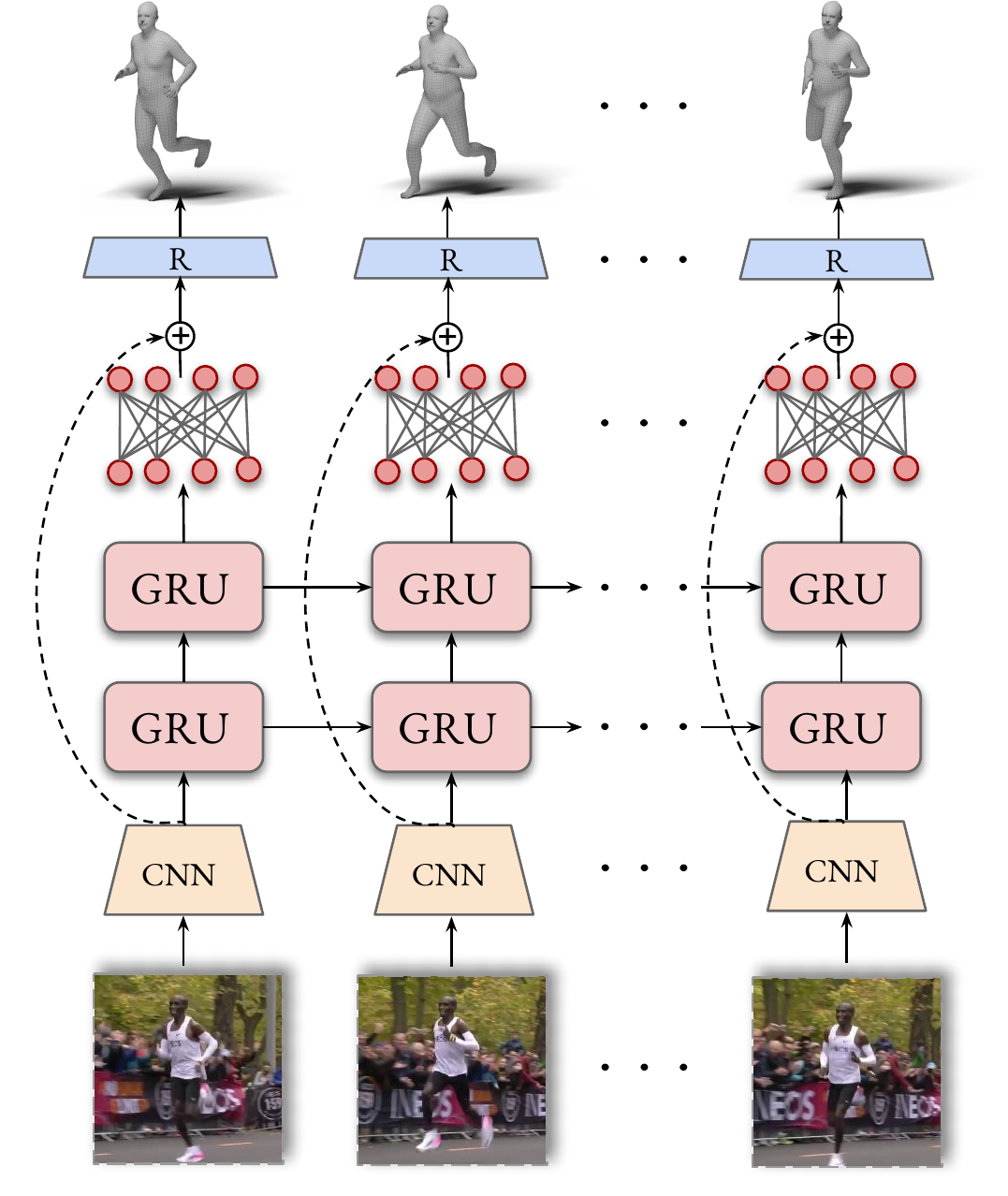}
	\caption{\textbf{Pose generator}, $\generator$, architecture used in our experiments. It takes a sequence of frames as input and outputs a vector $\in \mathbb{R}^{85}$.}
	\label{fig:generator}
\end{figure}

\paragraph{Pose Generator.} The architecture is depicted in Figure~\ref{fig:generator}. After feature extraction using ResNet50, we use a 2-layer GRU network followed by a linear projection layer. The pose and shape parameters are then estimated by a SMPL parameter regressor. We employ a residual connection to assist the network during training. The SMPL parameter regressor is initialized with the pre-trained weights from HMR~\cite{kanazawa_hmr,SPIN:ICCV:2019}. We decrease the learning rate if the reconstruction does not improve for more than 5 epochs.

\paragraph{Motion Discriminator.} We employ 2 GRU layers with a hidden size of 1024. For our most accurate results we used a self-attention mechanism with 2 MLP layers, each with 1024 neurons, and a dropout rate of $0.1$. For the ablation experiments we keep the same parameters changing the number of neurons and the number of MLP layers only. 

\paragraph{Loss.} We use different weight coefficients for each term in the loss function. The 2D and 3D keypoint loss coefficients are $\lambda_{2D} and \lambda_{3D}=300$ respectively, while $\lambda_{\beta} = 0.06$, $\lambda_{\theta}=60$. We set the motion discriminator adversarial loss term, $L_{adv}$ as $\lambda_{L_{adv}}=2$. We use 2 GRU layers with a hidden dimension size of $1024$.

\subsection{Datasets}
Below is a detailed summary of the different datasets we use for training and testing.

\vspace{0.1in}
\noindent{\bf MPI-INF-3DHP}~\cite{mpiiinf3dhp_mono-2017} is a multi-view, mostly indoor, dataset captured using a markerless motion capture system. We use the  training set proposed by the authors, which consists of 8 subjects and 16 videos per subject. We evaluate on the official test set. 

\vspace{0.1in}
\noindent{\bf Human3.6M}~\cite{ionescu_h36m} contains 15 action sequences of several individuals, captured in a controlled environment. There are 1.5 million training images with 3D annotations. We use SMPL parameters computed using MoSh~\cite{Mosh} for training. Following  previous work, our model is trained on 5 subjects (S1, S5, S6, S7, S8) and tested on the other 2 subjects (S9, S11). We subsampled the dataset to 25 frames per second for training.

\vspace{0.1in}
\noindent{\bf 3DPW}~\cite{vonMarcard2018_3dpw} is an in-the-wild 3D dataset that captures SMPL body pose using IMU sensors and hand-held cameras. It contains 60 videos (24 train, 24 test,  12  validation) of several outdoor and indoor activities. We use it both for evaluation and training.

\vspace{0.1in}
\noindent{\bf PennAction}~\cite{pennaction} dataset contains 2326 video sequences of 15 different actions and 2D human keypoint annotations for each sequence. The sequence annotations include the class label, human body joints (both 2D locations and visibility), 2D bounding boxes, and training/testing labels. 

\vspace{0.1in}
\noindent{\bf InstaVariety}~\cite{kanazawa_temporal_hmr} is a recently curated dataset using Instagram videos with particular action hashtags. It contains 2D annotations for about 24 hours of video. The 2D annotations were extracted using OpenPose~\cite{cao2018openpose} and Detect and Track~\cite{girdhar2018detecttrack} in the case of multi-person scenes.

\vspace{0.1in}
\noindent{\bf PoseTrack}~\cite{PoseTrack} is a benchmark for multi-person pose estimation and tracking in videos. It contains 1337 videos, split into 792, 170 and 375 videos for training, validation and testing respectively. In the training split, 30 frames in the center of the video are annotated. For validation and test sets, besides the aforementioned 30 frames, every fourth frame is also annotated. The annotations include 15 body keypoint locations, a unique person ID, and a head and a person bounding box for each person instance in each video. We use PoseTrack during training.

\subsection{Evaluation}
Here we describe the evaluation metrics and procedures we used in our experiments. 
For direct comparison we used the exact same setup as in ~\cite{SPIN:ICCV:2019}. Our best results are achieved with a model that includes the 3DPW training dataset in our training loop. Even without 3DPW training data, our method is more accurate than the previous SOTA.
For consistency with prior work, we use the Human3.6M training set when evaluating on the Human3.6M test set. 
In our best performing model on 3DPW, we do not use Human3.6M training data.
We observe, however, that strong performance on the Human3.6M does not translate to strong in-the-wild pose estimation.

\paragraph{Metrics.} We use standard evaluation metrics for each respective dataset. First, we report the widely used MPJPE (mean per joint position error), which is calculated as the mean of the Euclidean distances between the ground-truth and the predicted joint positions after centering the pelvis joint on the ground truth location (as is common practice). Also we report PA-MPJPE (Procrustes Aligned MPJPE), which is calculated similarly to MPJPE but after a rigid alignment of the predicted pose to the and ground-truth pose.
Furthermore, we calculate Per-Vertex-Error (PVE), which is denoted by the Euclidean distance between the ground truth and predicted mesh vertices that are the output of SMPL layer. 
We also use the Percentage of Correct Keypoints metric (PCK)~\cite{sapp2013modec}. The PCK counts as correct the cases where the Euclidean distance between the actual and predicted joint positions is below a predefined threshold.
Finally, we report acceleration error, which was reported in~\cite{kanazawa_temporal_hmr}. Acceleration error is the mean difference between ground-truth and predicted 3D acceleration for every joint ($mm/s^2$).

\end{document}